\documentclass[10pt,twocolumn,letterpaper]{article}

\usepackage{iccv}              

\definecolor{iccvblue}{rgb}{0.21,0.49,0.74}
\usepackage[pagebackref,breaklinks,colorlinks,allcolors=iccvblue]{hyperref}

\usepackage{tikz}
\usepackage{graphicx}

\usepackage{pifont}
\usepackage{algorithm}    
\usepackage{algpseudocode} 
\usepackage{amssymb}       
\usepackage{bm}            
\usepackage{array}
\usepackage{multirow}
\usepackage[table]{xcolor}
\usepackage{tcolorbox}
\usepackage{booktabs}
\usepackage{textcomp}
\usepackage{caption}
\usepackage{subcaption}  

\def\paperID{*****} 
\def\confName{ICCV}
\def\confYear{2025}

\title{Learning Hyperspectral Images with Curated Text Prompts for Efficient Multimodal Alignment}

\author{
Abhiroop Chatterjee, Susmita Ghosh \\
Jadavpur University \\
{\tt\small \{abhiroopchat1998, susmitaghoshju\}@gmail.com}
}

\makeatletter
\providecommand{\sf@counterlist}{}
\makeatother

\begin{document}

\maketitle
\begin{abstract}
As data requirements continue to grow, efficient learning increasingly depends on the curation and distillation of high-value data rather than brute-force scaling of model sizes. In the case of a hyperspectral image (HSI), the challenge is amplified by the high-dimensional 3D voxel structure, where each spatial location is associated with hundreds of contiguous spectral channels. While vision and language models have been optimized effectively for natural image or text tasks, their cross-modal alignment in the hyperspectral domain remains an open and underexplored problem. In this article, we make an attempt to optimize a Vision–Language Model (VLM) for hyperspectral scene understanding by exploiting a CLIP-style contrastive training framework. Our framework maps voxel-level embeddings from a vision backbone onto the latent space of a frozen large embedding model (LEM), where a trainable probe aligns vision features with the model’s textual token representations. The two modalities are aligned via a contrastive loss restricted to a curated set of hard (closest wrong classes) and semi-hard (random distractors) negatives, along with positive pairs. To further enhance alignment,    descriptive prompts that encode class semantics are introduced and act as structured anchors for the HSI embeddings. It is seen that the proposed method updates only \textbf{0.07\%} of the total parameters, yet yields state-of-the-art performance. For example, on \textit{Indian Pines (IP)} the model produces better results over unimodal and multimodal baselines by +0.92 Overall Accuracy (OA) and +1.60 Kappa ~$ (\kappa$), while on \textit{Pavia University (PU)} data it provides gains of +0.69 OA and +0.90~$\kappa$. Moreover, this is achieved with the set of parameters, nearly \textbf{50$\times$} smaller than DCTN and \textbf{90$\times$} smaller than SS-TMNet.
\end{abstract}    
\section{Introduction}

Hyperspectral imaging (HSI) \cite{HYPER1, GRSS} captures rich spatial--spectral information across hundreds of contiguous narrow bands, and facilitates fine-grained analysis of material properties and scene characteristics. Unlike RGB or multispectral imagery, which provide only a handful of broad channels, HSI encodes detailed spectral signatures that can be used to distinguish between objects that appear visually identical. This capability has made hyperspectral methods indispensable in domains \cite{HSI-review,HSI5,HSIMEDICINE,HSIMODERN} such as \textit{remote sensing}, \textit{environmental surveillance}, \textit{defense and security}, and \textit{biomedical imaging}.  

At the same time, this high-dimensional 3D data structure introduces fundamental challenges for \textit{representation learning} \cite{bengio2013representation}. Spectral redundancy, strong inter-band correlation, and the curse of dimensionality complicate feature extraction, while limited labeled datasets amplify the risk of overfitting. Models must also preserve the \textbf{spatial context} of scenes, critical for interpreting patterns such as vegetation distribution or tumor boundaries, and at the same time, exploit the fine-grained spectral features that distinguish materials. Moreover, HSI tasks differ from conventional vision problems in several respects. We are listing some of them below:  
\begin{enumerate}
    \item \textbf{Spectral precision} -- Success often depends on identifying minute spectral differences invisible to RGB-based systems.  
    \item \textbf{Data limitations} -- High acquisition cost and complex sensor setups result in smaller, domain-specific datasets compared to large-scale benchmarks such as ImageNet.  
    \item \textbf{Evaluation protocol} --  \textbf{Labeling hyperspectral pixels is extremely difficult and expensive}, the evaluation setup \cite{DCTNINGARSS} is often \textbf{reversed compared to standard vision tasks}. Instead of training on large annotated sets and testing on small subsets, HSI models are commonly trained with only \textbf{10\% labeled training data} and evaluated on the remaining \textbf{90\% test data}. This type of protocol reflects the practical reality of scarce supervision and shows the need for efficient learning methods under limited label scenarios.  
    
\end{enumerate}

These differences raise the need to build newer methods to progress hyperspectral analysis. Recent advances have explored diverse vision architectures \cite{CNN,he2016deep,krizhevsky2012imagenet,vaswani2017attention,dosovitskiy2020image} to address these challenges. We highlight the \textbf{state-of-the-art methods} \cite{2DCNNINGARSS,3DCNNINGARSS,DCTNINGARSS,SSTMINGARSS,SSFTT,MORPH,HYBRIDSN} along with \textbf{established world models} \cite{2DCNN,3DCNN,dosovitskiy2020image,CLIP} below:

\textbf{Literature Review}. Early HSI models relied on convolutional approaches. \textbf{2D-CNN}~\cite{2DCNNINGARSS} extracts spatial features via stacked 2-D convolutions, while \textbf{3D-CNN}~\cite{3DCNNINGARSS} jointly models spatial–spectral information. Hybrid models like \textbf{HybridSN}~\cite{HYBRIDSN} combine 3D and 2D convolutions, and capture both spatial–spectral and spatial features. Transformer-based methods, including \textbf{ViT}~\cite{dosovitskiy2020image,Han2022SurveyVisionTransformer}, \textbf{SSFTT}~\cite{SSFTT}, and \textbf{morphFormer}~\cite{MORPH}, exploit attention to model long-range dependencies. Other architectures such as \textbf{HiT}~\cite{HITINGARSS} and \textbf{SS-TMNet}~\cite{SSTMINGARSS} use spectral A3D convolutions and multiscale spatial–spectral attention. On the other hand, dual-branch networks like \textbf{DCTN}~\cite{DCTNINGARSS}, combine CNNs for local features with transformers for global spectral modeling to achieve state-of-the-art performance. Contrastive vision–language frameworks \cite{VLM,li2025survey} such as \textbf{CLIP}~\cite{CLIP} provide transferable cross-modal embeddings, but are designed for natural 2D images and overlook volumetric spatial-spectral structures in HSI. Several remote sensing applications often require understanding complex 3D patterns, and this motivates us to explore the multimodal alignment \cite{CLIP} in the hyperspectral domain.

\begin{figure*}[t]
    \centering
    \subfloat[]{%
        \begin{tikzpicture}
            \node[
                draw=black,
                dotted,
                line width=0.8pt,
                rounded corners=8pt,
                inner sep=2pt
            ] {
                \includegraphics[width=0.71\linewidth, height=0.32\textheight]{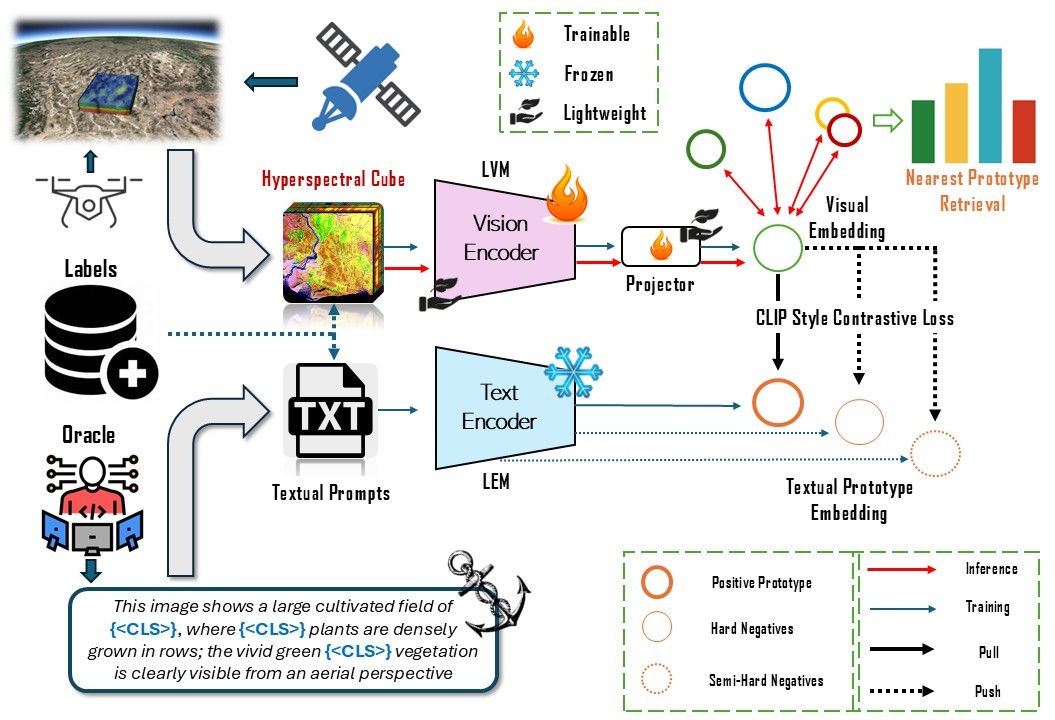}
            };
        \end{tikzpicture}
        \label{fig:vlm_flow}
    }
    \hfill
    \subfloat[]{%
        \begin{tikzpicture}
           \node[
                draw=black,
                dotted,
                line width=0.8pt,
                rounded corners=8pt,
                inner sep=2pt
            ] {
                \includegraphics[width=0.25\linewidth, height=0.32\textheight]{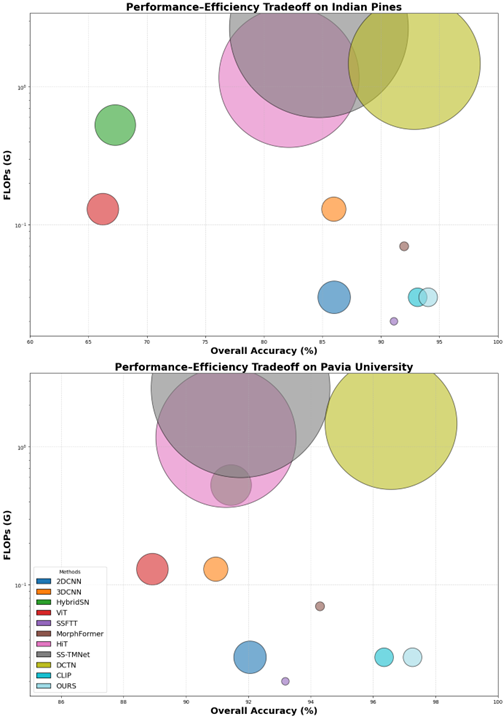}
            };
        \end{tikzpicture}
        \label{fig:performance_efficiency}
    }

    \caption{\textbf{Graphical snapshot of our approach}: 
    (a) Vision–Language Model for hyperspectral scene understanding. 
    (b) Performance–Efficiency tradeoff across different methods on IP (up) and PU (down); the size of a circle is proportional to the total parameters (MB). The x-axis denotes the overall accuracy in (\%)  and the y-axis represents FLOPs in G. The legend is common for both datasets.}
    \label{fig:overview}
\end{figure*}

Here, we present a \textit{Vision Language Model} (Figure \ref{fig:vlm_flow}), for \textit{hyperspectral scene understanding}, that aligns a ViT backbone with a frozen large embedding model (LEM) \cite{LLM,LLM2,LLM3} through contrastive learning. To boost discriminability, we employ \emph{descriptive prompts} encoding class semantics and \emph{informative negatives} to counter hard distractors. It is seen that training around 0.07\% of parameters, our method achieves SOTA on various HSI benchmarks. A parameter efficiency snapshot is given in Figure \ref{fig:performance_efficiency}.

The article is organized as follows: Section~\ref{sec:methodology} details the proposed methodology, while Section~\ref{exp} describes the experimental setup. Section~\ref{res} presents the results with analysis, and Section~\ref{con} concludes with an outlook and future directions.

\section{Methodology}
\label{sec:methodology}

Let $\mathcal{X} \subset \mathbb{R}^{H \times W \times D}$ denote hyperspectral patches and $\mathcal{Y} = \{1, \dots, C\}$ the set of $C$ classes. Our goal is to learn a visual embedding function, $f_\theta: \mathcal{X} \to \mathbb{R}^d$, that maps semantically similar patches closer in the latent space. To provide semantic guidance, we employ a LEM, $
g_\phi: \mathcal{T} \to \mathbb{R}^d,
$ 
which maps textual prompts $\mathcal{T}$ into the shared latent space. Both visual and textual embeddings are $\ell_2$-normalized:
\begin{equation}
\mathbf{z}_x = \frac{f_\theta(x)}{\|f_\theta(x)\|_2}, \quad 
\mathbf{z}_t = \frac{g_\phi(t)}{\|g_\phi(t)\|_2}.
\end{equation}
This formulation induces a Riemannian geometry \cite{shao2018riemannian} on the unit hypersphere $\mathbb{S}^{d-1}$, and allows the similarity between visual and textual embeddings to be measured via the cosine of the geodesic angle.

\textbf{Prompt Engineering and LEM Embeddings}. Each class $C$ is assigned a single descriptive prompt $t_C$ that narrates the distinguishing visual and semantic characteristics of hyperspectral patches, such as crop type, cultivation method, vegetation density, or aerial perspective. These prompts are designed from domain knowledge and are made to be informative, discriminative, and non-redundant across classes, so that the resulting embeddings are maximally separated in semantic space. The LEM, $g_\phi$, maps each prompt to a fixed embedding. During training, the LEM is linear-probed, and the textual embeddings are kept fixed. These embeddings serve as \textit{semantic anchors} for the CLIP-style \cite{CLIP} contrastive objective, and guide the visual embeddings to align with the corresponding class semantics. The prompts are designed in this manner:

\begin{tcolorbox}[colback=blue!5!white, colframe=blue!50!black, boxrule=0.4mm]
\textbf{\ding{52} \small Descriptive Prompt Template}. 
This image shows a large cultivated field of \textcolor{blue}{\textbf{\{\texttt{<CLS>}\}}}, 
where \textcolor{blue}{\textbf{\{\texttt{<CLS>}\}}} plants are densely grown in rows; 
the vivid green \textcolor{blue}{\textbf{\{\texttt{<CLS>}\}}} vegetation is clearly visible 
from an aerial perspective.
\end{tcolorbox}

Other classes use similarly structured prompts with \texttt{<CLS>} as the class placeholder. For the vision backbone, we employ \textit{Masked Vision Transformer} \cite{dosovitskiy2020image, he2022masked,ICCVAGMVIT}, that we train end-to-end on hyperspectral patches using the proposed contrastive objective.

\textbf{Contrastive Loss with Hard and Semi-Hard Negatives}. Let $\mathbf{z}_i = \mathbf{z}_{x_i} \in \mathbb{R}^d$ denote the $\ell_2$-normalized embedding of the $i^{th}$ hyperspectral patch produced by the vision encoder, and let $\mathbf{p}_j = \mathbf{z}_{t_j} \in \mathbb{R}^d$ denote the $\ell_2$-normalized textual prototype embedding for class $j$.

We now define a scaled cosine similarity between an image embedding and a class prototype as:
\begin{equation}
s_{ij} = \tau \,\mathbf{z}_i^\top \mathbf{p}_j, 
\quad \tau = e^{\text{logit\_scale}}, 
\quad i \in \mathcal{B}, \, j \in \{1, \dots, C\},
\end{equation}

where $\tau > 0$ is a learnable temperature scaling the distribution, and $\mathcal{B}$ denotes the training batch.

\paragraph{Positive logit}  
For patch $i$ with ground-truth label $y_i$, the positive logit is: $
s_i^+ = s_{i y_i},
$
representing similarity to the correct class prototype.

\paragraph{Hard negatives}  
Let $k_h$ denote the number of top-hard negatives. For patch $i$, the indices of the top-$k_h$ most confusing incorrect classes are:
\begin{equation}
H_i = \text{Top-}k_h\Big(\{ s_{ij} \mid j \neq y_i \}\Big),
\end{equation}
and the corresponding logits are $
s_i^{\text{hard}} = \{ s_{ij} \mid j \in H_i \}.
$

\paragraph{Semi-hard negatives}  
Let $k_s$ denote the number of semi-hard negatives. We randomly sample $k_s$ number of negatives from the remaining classes:
\begin{equation}
SH_i \subset \{1, \dots, C\} \setminus (\{y_i\} \cup H_i), \quad |SH_i| = k_s,
\end{equation}
with logits: $
s_i^{\text{semi}} = \{ s_{ij} \mid j \in SH_i \}.
$

\paragraph{Combined logits}  
We then concatenate the positive, hard, and semi-hard negatives to form the final logit vector:
\begin{equation}
\mathbf{s}_i = [\, s_i^+, \, s_i^{\text{hard}}, \, s_i^{\text{semi}} \,] \in \mathbb{R}^{1 + k_h + k_s}.
\end{equation}
The positive logit is always the first entry, followed by the hardest and then the semi-hard negatives.

\paragraph{Loss computation}  
The cross-entropy loss over the combined logits is:
\begin{equation}
\mathcal{L}_i \;=\; 
- \log 
\frac{e^{\,s_i^+}}
{e^{\,s_i^+} \;+\; 
 \sum_{j \in H_i} e^{\,s_{ij}} \;+\; 
 \sum_{j \in SH_i} e^{\,s_{ij}} }.
\end{equation}

The batch-level loss is computed as:
\begin{equation}
\mathcal{L} = \frac{1}{|\mathcal{B}|} \sum_{i \in \mathcal{B}} \mathcal{L}_i.
\end{equation}

This objective aligns each visual embedding $\mathbf{z}_i$ with its textual prototype $\mathbf{p}_{j}$, while hard and semi-hard negatives enforce fine-grained discrimination on the hypersphere $\mathbb{S}^{d-1}$.

\begin{figure*}[ht]
    \centering
    \includegraphics[width=1\linewidth, height=0.28\textheight]{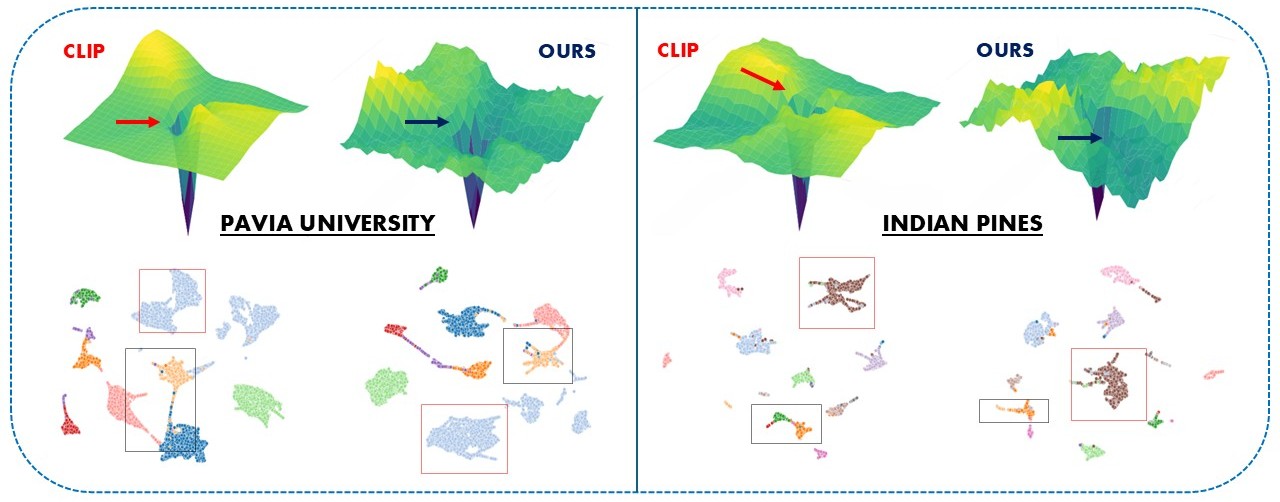}
   \vspace{-0.2in}
    \caption{Loss landscape (top) and UMAP projections (bottom). From left to right: PU and IP datasets. Our model shows a more structured, generalizable landscape and neural-collapse-like embeddings, unlike the smoother but overlapping CLIP features. For instance, CLIP sometimes displays a butterfly-like structure for classes with larger samples, whereas our method exhibits a nearly circular embedding with compact intra-class relationships. Also, the inter-class separation observed is more in our case. Please zoom for better clarity.}
    \label{fig:umap_pavia_ip}
\end{figure*}

\begin{tcolorbox}[colback=green!5!white, colframe=green!50!black, boxrule=0.5mm]

\textbf{\ding{52} Distractor-Aware Contrastive Alignment}.
The vision and text embeddings are aligned via a contrastive loss on positive pairs and carefully curated hard and semi-hard negatives. Hard negatives sharpen class boundaries, while semi-hard negatives introduce variability. This optimization jointly enhances training efficiency (Table \ref{tab:methods_combined_backbone}, \ref{tab:training_samples_indianpines_transpose_oa}, \ref{tab:prompt_vs_vision}, \ref{tab:method_backbone_comparison}) and embedding separability (Figure \ref{fig:umap_pavia_ip}) at the same time.

\end{tcolorbox}

\textbf{Inference Procedure}. During inference, a hyperspectral patch $x \in \mathcal{X}$ is passed through the trained vision backbone $f_\theta$ to obtain its $\ell_2$-normalized embedding $\mathbf{z}_x \in \mathbb{R}^d$. This embedding is then compared against the fixed set of textual prototypes $\{\mathbf{p}_j\}_{j=1}^C$, derived from the class-specific prompts, using cosine similarity. The predicted class label is determined by \textit{nearest-prototype retrieval}: $
\hat{y} = \arg\max_{j \in \{1, \dots, C\}} \; \mathbf{z}_x^\top \mathbf{p}_j.
$
This way, the classification at inference time reduces to a similarity search in the shared cross-modal embedding space, with \textit{no additional trainable parameters required}.

\definecolor{gold}{RGB}{255, 215, 0}
\definecolor{silver}{RGB}{192, 192, 192}
\definecolor{cyanlight}{RGB}{210, 245, 255}
\definecolor{cnncolor}{RGB}{240, 240, 240}
\definecolor{transformercolor}{RGB}{255, 250, 205}
\definecolor{multimodalcolor}{RGB}{220, 245, 220}
\begin{table*}
\scriptsize
\centering
\caption{Comparison with other SOTA methods on various HSI datasets and their computational efficiency. \textbf{Bold}, \textcolor{blue}{blue}, and \textcolor{red}{red} represent the highest/lowest, second, and third performance/efficiency. M-CLS represents multi CLS prompts.}
\label{tab:methods_combined_backbone}
\begin{tabular}{l l l ccc ccc ccc}
\toprule
\textbf{Category} & \textbf{Backbone} & \textbf{Methods} & \multicolumn{3}{c}{\textbf{Indian Pines}} & \multicolumn{3}{c}{\textbf{Pavia University}} & \textbf{FLOPs (G)} & \textbf{Params (MB)} & \textbf{Latency Time (s)} \\
\cmidrule(lr){4-6} \cmidrule(lr){7-9}
& & & \textbf{OA (\%)} & \textbf{$\kappa$} & & \textbf{OA (\%)} & \textbf{$\kappa$} & & & & \\
\midrule

\rowcolor{cnncolor}
\textbf{Unimodal} & 2D CNN & 2DCNN \citep{2DCNNINGARSS} & 85.98 & 84.22 & & 92.05 & 89.84 & & \textcolor{blue}{0.03} & 2.76 & \textcolor{blue}{2.04} \\
\rowcolor{cnncolor}
& 3D CNN & 3DCNN \citep{3DCNNINGARSS} & 85.95 & 83.91 & & 90.95 & 88.44 & & 0.13 & 1.54 & 6.54 \\
\rowcolor{cnncolor}
& 2D + 3D CNN & HybridSN \citep{HYBRIDSN} & 67.26 & 62.21 & & 91.44 & 89.06 & & 0.53 & 4.32 & 7.78 \\
\midrule

\rowcolor{transformercolor}
\textbf{Unimodal} & Vision Transformer & ViT \citep{dosovitskiy2020image} & 66.21 & 61.65 & & 88.92 & 85.81 & & 0.13 & 2.61 & \textcolor{red}{5.31} \\
\rowcolor{transformercolor}
& Token Transformer & SSFTT \citep{SSFTT} & 91.11 & 89.94 & & 93.18 & 91.25 & & \textbf{0.02} & \textbf{0.15} & \textbf{1.18} \\
\rowcolor{transformercolor}
& Morph-ATT Transformer & MorphFormer \citep{MORPH} & 91.98 & 90.91 & & 94.29 & 92.63 & & \textcolor{red}{0.07} & \textcolor{blue}{0.21} & 9.19 \\
\rowcolor{transformercolor}
& Hyperspectral Transformer & HiT \citep{HITINGARSS} & 82.13 & 79.77 & & 91.28 & 88.85 & & 1.17 & 51.23 & 11.68 \\
\rowcolor{transformercolor}
& Multi-Scale Transformer & SS-TMNet \citep{SSTMINGARSS} & 84.67 & 82.66 & & 91.74 & 89.44 & & 2.67 & 83.33 & 31.12 \\
\rowcolor{transformercolor}
& CNN + Transformer & DCTN \citep{DCTNINGARSS} & \textcolor{red}{92.85} & \textcolor{red}{91.87} & & \textcolor{blue}{96.57} & \textcolor{blue}{95.49} & & 1.48 & 45.32 & 20.69 \\
\midrule

\rowcolor{multimodalcolor}
\textbf{Multimodal} & Vision Language Model & M-CLS CLIP \cite{CLIP} & \textcolor{blue}{93.11} & \textcolor{blue}{91.94} & & \textcolor{red}{96.35} & \textcolor{red}{94.50} & & \textcolor{blue}{0.03} & \textcolor{red}{0.91} & 13.84 \\
\midrule

\rowcolor{cyan!20}
\textbf{Multimodal} & LEM + LVM & \textbf{OURS} & \textbf{94.03} & \textbf{93.54} & & \textbf{97.26} & \textbf{96.39} & & \textcolor{blue}{0.03} & \textcolor{red}{0.91} & 14.19 \\
\midrule

\rowcolor{orange!20}
&  & $\Delta$ & \textbf{0.92} & \textbf{1.60} & & \textbf{0.69} & \textbf{0.90} & &  &  &  \\
\bottomrule
\end{tabular}
\end{table*}

\section{Experimental Setups}
\label{exp}

This section outlines the experimental setup used to train on two benchmark hyperspectral datasets~\citep{hyperspectral}.

\textbf{Datasets Used.} We evaluate on two widely-used benchmark hyperspectral datasets: \textit{Indian Pines} \cite{hyperspectral} and \textit{Pavia University} \cite{hyperspectral}.  The \textit{Indian Pines (IP)} dataset was collected by the AVIRIS sensor over agricultural fields in Northwestern Indiana. It has a spatial size of $145 \times 145$ pixels and originally contains 220 spectral bands ranging from 0.4 to 2.5 $\mu$m. After removing water absorption bands, 200 bands are retained. The dataset is annotated into 16 land-cover classes, most of which are related to different types of crops (e.g., corn, soybeans, alfalfa), along with a few classes corresponding to natural vegetation and man-made structures. IP is particularly challenging due to its \textbf{high class imbalance}, \textbf{small sample sizes}, and the presence of spectrally similar vegetation classes. Total number of samples in the IP dataset is 10,249.  

The \textit{Pavia University (PU)} dataset was acquired by the ROSIS sensor over the University of Pavia, Italy. It has a larger spatial coverage of $610 \times 340$ pixels with 115 spectral bands, of which 103 remain after discarding noisy channels. PU contains 9 land-cover classes, including urban features (e.g., asphalt, bitumen, bricks, shadows), vegetation, and bare soils. Compared to IP, PU offers \textbf{higher spatial resolution (1.3 m per pixel)} and more spatially coherent regions, which makes it suitable for evaluating the ability of models to capture both spectral signatures and spatial context in structured urban environments.  The number of samples in the PU dataset is 42,776, and it is the larger of the two.   

\textbf{Data Pre-Processing Pipeline Used.} Given the high dimensionality and redundancy in hyperspectral data, we apply a zero padding to preserve the spatial structure at image borders and perform a Principal Component Analysis (PCA) \cite{PCA} to reduce the spectral dimension to 25 principal components.

\textbf{Vision Language Model Summary}. The designed VLM has $335.3$M parameters, with $\sim335$M frozen in the text encoder and only $\sim240$K trainable. Of these, the vision encoder contributes $174$K parameters, the projection head $65.6$K, and a single logit scale parameter. Each $3 \times 3$ patch is projected into a 64-dimensional embedding and processed through 6 transformer layers with 16 self-attention heads and an MLP dimension of 64. This compact vision branch aligns with the frozen LEM (\texttt{BAAI/bge-large-en-v1.5} \cite{bge_large_en_v1_5}), which provides stable 1024-D textual embeddings for descriptive prompts.

\textbf{Fine-tuning Configuration.} The model is trained in a vision–language alignment setup with both hard and semi-hard negatives. On Indian Pines, we train for $50$ epochs with batch size $32$, while on Pavia University we train for $25$ epochs with batch size $128$. Following the DCTN \cite{DCTNINGARSS} protocol, $10\%$ of labeled data is used for training and $90\%$ for testing.   Adam optimizer is used with a learning rate of $1 \times 10^{-3}$. Contrastive training uses $k_h=4$ hard and $k_s=4$ semi-hard negatives, with a CLIP-style loss restricted to $\{s^+, s^{\text{hard}}, s^{\text{semi}}\}$ (Sec.~\ref{sec:methodology}). All the experiments were simulated for four times on an NVIDIA A100 GPU.

\section{Analysis of Results}
\label{res}

In this section, we compare the proposed VLM with a set of state-of-the-art approaches and observe that it delivers performance on par with the best existing methods and yields high accuracy at a very low computational cost. 

\textbf{Performance Evaluation Metrics.} For evaluation, we report both \textbf{Overall Accuracy (OA)} and \textbf{Cohen’s Kappa coefficient ($\kappa$)}. Overall Accuracy (OA) measures the proportion of correctly classified samples across all classes, offering an indicator of classification performance. The $\kappa$ coefficient, on the other hand, accounts for the agreement occurring by random chance and provides a chance-corrected measure of reliability.

\textbf{Comparison with SOTA Approaches.}  
Table~\ref{tab:methods_combined_backbone} ellucidates the comparative performance of the proposed VLM against existing HSI classifiers. CNN-based unimodal baselines such as 2D-CNN \cite{2DCNNINGARSS}, 3D-CNN \cite{3DCNNINGARSS}, and HybridSN \cite{HYBRIDSN} deliver moderate accuracy. This is primarily due to their limited ability to capture long-range spectral–spatial dependencies. Transformer-based methods (SSFTT \cite{SSFTT}, MorphFormer \cite{MORPH}, DCTN \cite{DCTNINGARSS}) demonstrate clear gains and emphasize the advantage of global attention in modeling hyperspectral features. Our multimodal VLM consistently surpasses these unimodal architectures and shows the strength of cross-modal alignment. Among all the methods, DCTN \cite{DCTNINGARSS} stands out as a hybrid architecture that combines CNNs with transformers, thus integrating both local spatial cues and long-range global dependencies. This dual design makes it the strongest vision-only backbone, capable of expressive feature learning. However, it is worth noting that optimizing unimodal vision models is often more straightforward and sometimes yields highly expressive representations, since they are not constrained by the alignment challenges posed by an additional modality. Nevertheless, even the strongest unimodal transformer backbone (DCTN) is sometimes outperformed by multimodal approaches such as CLIP \cite{CLIP}, and this validates the importance of incorporating meaningful textual priors. CLIP \cite{CLIP} performs competitively versus DCTN \cite{DCTNINGARSS} while being lighter. We observe that our method surpasses all unimodal and multimodal SOTAs on Indian Pines (+0.92 OA, +1.60 $\kappa$) and Pavia (+0.69 OA, +0.90 $\kappa$), and validates that descriptive prompts and informative negatives enable stronger cross-modal grounding.

\textbf{Parameter and Latency Analysis}.
Table~\ref{tab:methods_combined_backbone} highlights the trade-off between cost and performance. CNN baselines \cite{2DCNNINGARSS,3DCNNINGARSS} are lightweight and fast but limited in accuracy. Transformer models such as SSFTT \cite{SSFTT} offer efficiency with fewer FLOPs and parameters, while heavier ones like MorphFormer \cite{MORPH} and DCTN \cite{DCTNINGARSS} demand more resources. Multimodal methods, including CLIP \cite{CLIP} and our method, add cross-modal alignment overhead. Yet, the proposed model achieves higher accuracy than CLIP without extra FLOPs or memory. We also notice an optimal trade-off between state-of-the-art accuracy with minimal cost and low inference time.

\textbf{Ablation on Hard vs. Semi-Hard Negatives.} 
Table~\ref{tab:ablation_indianpines} presents the impact of different negative sampling strategies on our contrastive alignment framework. Removing hard negatives results in a significant performance drop to 90.16\% OA, and shows their importance in disentangling closely related spectral classes. Excluding semi-hard negatives leads to a higher OA of 93.44\%, but still underperforms compared to the full design. Incorporating both hard and semi-hard negatives achieves the best result of 94.03\% OA. From this observation in Table~\ref{tab:ablation_indianpines}, we can infer that \textbf{hard negatives} drive the model to resolve inter-class ambiguities by pushing apart spectrally similar but semantically different categories (e.g., different crop types in Indian Pines). Meanwhile, \textbf{semi-hard negatives} act as regularizers and refine the decision boundaries. This helps in preventing the model from collapsing features of borderline or underrepresented classes. Together, their joint presence achieves both separability and robust generalization (Figure \ref{fig:umap_pavia_ip}).

\definecolor{cyanlight}{RGB}{210, 245, 255}
\definecolor{orangelight}{RGB}{255, 235, 205}

\begin{table}
\small
\centering
\caption{Ablation study of \textbf{varying loss components} on IP.}
\label{tab:ablation_indianpines}
\begin{tabular}{l c}
\toprule
\textbf{Variant} & \textbf{OA (\%)} \\
\midrule

\rowcolor{cyanlight}
\ding{55} without Hard & 90.16  \\
\rowcolor{orangelight}
\ding{55} without Semi-Hard & 93.44  \\

\midrule
\rowcolor{orangelight}
\ding{52} Full Mode (Hard + Semi-Hard) & \textbf{94.03} \\
\bottomrule
\end{tabular}
\end{table}

\definecolor{cyanlight}{RGB}{210, 245, 255}
\definecolor{orangelight}{RGB}{255, 235, 205}

\begin{table}
\small
\centering
\caption{OA (\%) comparison of {proposed technique} with {DCTN} \cite{DCTNINGARSS} on Indian Pines with varying training sample percentages.}
\label{tab:training_samples_indianpines_transpose_oa}
\begin{tabular}{l c c c c c}
\toprule
\textbf{Method} & 10\% & 20\% & 30\% & 40\% & 50\% \\
\midrule
\rowcolor{cyanlight}
DCTN \cite{DCTNINGARSS} & 92.85 & 95.37 & 95.81 & 96.01 & 96.10 \\
\midrule
\rowcolor{orangelight}
OURS & \textbf{94.03} & \textbf{97.68} & \textbf{98.57} & \textbf{98.97} & \textbf{99.02} \\
\bottomrule
\end{tabular}
\end{table}

\textbf{Training Data Sensitivity}.
Table~\ref{tab:training_samples_indianpines_transpose_oa} shows the sensitivity with varying supervision.  Despite being one of the strongest baselines, DCTN \cite{DCTNINGARSS} already achieves competitive overall accuracies (OA) on Indian Pines, and shows its effective integration of convolutional and transformer-based modeling. However, the proposed model consistently surpasses DCTN across all training sample percentages. The improvement is seen under scarce supervision (10\% training data), where our method yields a \textbf{+1.18\%} OA gain (94.03\% vs.\ 92.85\%). This advantage further grows at higher data availability, and reaches near-saturation with \textbf{99.02\%} OA at 50\% training data. This suggests that the proposed approach not only better exploits the limited labeled samples, which is a key challenge in hyperspectral learning, but also scales more effectively as training data increases. In contrast, the performance of DCTN plateaus earlier and indicates limitations in capturing informative and more discriminative spectral–spatial features. The steady margin maintained by our model corroborates its stronger representational efficiency across varying levels of supervision.

\textbf{Ablation Study on Varying Batch Size.}  
Table~\ref{tab:batch_size_ours} reports the effect of varying batch sizes on classification performance for Indian Pines. We observe a clear trade-off between representation quality and training stability. Small batch sizes (e.g., 4 or 8) provide sufficient gradient diversity but may suffer from noisier updates, and yield OA values of 93.19\% and 93.45\%, respectively. Increasing the batch size to 16 improves stability and achieves 93.54\%, while a batch size of 32 provides the best balance and results in the highest overall accuracy of \textbf{94.03\%}.

\textbf{Sensitivity to Prompt Representation.}  
The results in Table~\ref{tab:text_prompt_comparison_scientific} depict the role of prompt design in evolving performance. When the model is guided by label-only prompts, it receives little more than a name tag and offers minimal semantic context. Short-text prompts, while an improvement, resemble terse dictionary entries that hint at meaning but fail to capture the full richness of the scene. Contrary to that, our descriptive long-text prompts act more like well-structured narratives, and try to embed both distinction and context that guide the model to anchor visual features to meaningful linguistic constructs. \textit{It is much like how we get a deeper comprehension when immersed in a full passage rather than a single word; the model achieves stronger cross-modal alignment when exposed to richer descriptions}.

\definecolor{cyanlight}{RGB}{210, 245, 255}

\begin{table}
\scriptsize
\centering
\caption{ablation study of \textbf{varying batch sizes} on Indian Pines.}
\label{tab:batch_size_ours}
\begin{tabular}{l c c c c c c}
\toprule
\textbf{Batch Size} & 4 & 8 & 16 & 32 & 64 & 128 \\
\midrule
\rowcolor{cyanlight}
\textbf{OURS OA (\%)} & 93.19 & 93.45 & 93.54 & \textbf{94.03} & 93.34 & 91.12 \\
\bottomrule
\end{tabular}
\end{table}

\begin{tcolorbox}[colback=blue!5!white,colframe=blue!50!black,boxrule=0.4mm,left=2mm,right=2mm,top=1mm,bottom=1mm,overlay={
    \draw[blue!50!black,dashed,line width=0.4mm,rounded corners=5mm]
        (frame.south west) rectangle (frame.north east);
}]
\textbf{\textbf{\ding{52} But why should a few extra words make such a difference?}}  
Because in multimodal learning, every additional semantic pattern becomes a bridge 
that can tie abstract text to vision and more bridges mean stronger cross-modal alignment.
\end{tcolorbox}

\definecolor{cyanlight}{RGB}{210, 245, 255}
\begin{table}[H]
\small
\centering
\caption{Comparison of different \textbf{text prompt types} on the Indian Pines (IP) dataset. The results show how variations in prompt design influence classification performance.}
\label{tab:text_prompt_comparison_scientific}
\begin{tabular}{l c c c }
\toprule
\textbf{LEM Prompt Type} & \textbf{Label-only} & \textbf{Short Text} & \textbf{OURS} \\
\midrule
\rowcolor{cyanlight}
\textbf{OA (\%)} & 92.90 & 93.07 & \textbf{94.03}  \\
\bottomrule
\end{tabular}
\end{table}

\definecolor{bgecolor}{RGB}{240,248,255}      
\definecolor{m3color}{RGB}{255,250,205}       
\definecolor{e5color}{RGB}{220,245,220}       
\vspace{-0.2in}
\begin{table}[H]
\small
\centering
\caption{Effect of various \textbf{text-embedding backbones} on IP.}
\label{tab:llm_backbone_ablation}
\begin{tabular}{l l l c}
\toprule
\textbf{LEM Backbone} & \textbf{Type} & \textbf{Family} & \textbf{OA (\%)} \\
\midrule
\rowcolor{bgecolor}
BAAI/bge-large-en-v1.5 & English & BGE & \textbf{94.03} \\
\rowcolor{m3color}
BAAI/bge-M3 & Multilingual & BGE & 93.04 \\
\rowcolor{e5color}
E5-Large (multilingual) & Multilingual & E5 & 92.72 \\
\bottomrule
\end{tabular}
\end{table}

\textbf{Ablation Study on LEM Backbone Choice}. 
Table~\ref{tab:llm_backbone_ablation} compares different text-embedding backbones used. We observe that the \textit{English-only} BGE model (BAAI/bge-large-en-v1.5) \cite{bge_large_en_v1_5} achieves the highest performance, reaching \textbf{94.03\%} OA. This shows us that for hyperspectral classification tasks, where the label space is relatively small, fixed, and dominated by English terminology, a strong monolingual embedding model can provide highly discriminative representations.  In contrast, the multilingual variants, although more general-purpose, exhibit slightly lower performance: BGE-M3 achieves 93.04\% OA, while the E5-Large multilingual model reaches 92.72\% OA. This drop may be attributed to the fact that multilingual models spread their capacity across many languages and sacrifice specialization in English, where the task-specific labels reside.

\textbf{Analysis of Prompt-based vision language model with Vision-only Baseline}.  
Comparing a descriptive prompt–based vision language model against a vision-only baseline depicts the impact of textual priors in Table~\ref{tab:prompt_vs_vision}. On both IP and PU datasets, the prompt-based approach exceeds vision-only performance (94.03\% vs 91.52\%, 97.26\% vs 96.60\%). This gap illustrates that language-derived context enhances class separability even in high-dimensional hyperspectral spaces. Moreover, the enriched semantic guidance helps resolve ambiguous spectral signatures and improves the robustness of learned embeddings across varying scenes. Figure~\ref{fig:umap_pavia_ip} visually confirms that textual priors lead to more intra-class compactness and better inter-class margins, and also shows that textual information is not merely auxiliary but can be central in achieving high performance across datasets.

\textbf{Analysis of the Loss Landscape}. We analyze the training dynamics of our model with CLIP \cite{CLIP} in Figure \ref{fig:umap_pavia_ip}.  Loss landscape in CLIP is flat (from some directions) and smoother, while our model shows a \textit{\underline{slightly slanted (from numerous directions)}}, \textit{\underline{rugged landscape}} with a \textit{\underline{larger minimum diameter}}, and this suggests better generalization and reduced risk of getting trapped in saddle points. It is worth noting that while smooth, flat loss landscapes reduce sharp minima, but can give weaker directional optimization guidance. Excessively isotropic minima often yield Hessian spectra with compressed eigenvalue distributions, and suppress anisotropic curvature and informative descent directions. This spectral degeneracy weakens the signal-to-noise gradient signal and limits the optimizer's exploitation of the principal curvature subspaces.

\textbf{Qualitative Visuals of the Embeddings}. The Uniform Manifold Approximation and Projection (UMAP) \cite{UMAP} analysis show that our technique's embeddings exhibit a near neural-collapse-like structure \cite{papyan2020prevalence}, with tightly clustered class centroids, maximal inter-class separation, and minimal intra-class variance, contrasting with the more overlapping CLIP embeddings. Here, we attempt to analyze the UMAP projections from two complementary perspectives: (1) Latent geometry and (2) Latent separation. With respect to (1), the CLIP-induced embeddings show non-Euclidean dispersion patterns, with certain semantic manifolds exhibiting a butterfly-shaped bifurcation for PU. This effect is most conspicuous in the light-blue class (Figure \ref{fig:umap_pavia_ip}), where the embedding unfolds into two lobes joined by a narrow topological corridor. Similarly, the light-green class also shows anisotropic curvature and is characterized by elongated eigen-directions that expose distortions in the latent geometry. Contrary to that, our approach yields manifolds that are topologically closer to isotropic Gaussian distributions, looking like circular basins in the UMAP plane. We also observe that this isotropy persists even for the deep red-labeled PU class, which otherwise exhibits higher-order deformations under CLIP. A parallel observation holds for the IP dataset also, thus ensuring robustness across datasets.

With respect to (2) Latent separation, in the CLIP embeddings of the Indian Pines (IP) dataset, we observe a noticeable inter-class connectivity involving three distinct classes, and we highlight this with a black boundary in left Sub-Figure~\ref{fig:umap_pavia_ip} for clarity. This phenomenon reflects residual overlaps in the embedding space, where class-specific manifolds remain entangled rather than well-separated. However, our approach reduces this effect. The same region that shows entanglement under CLIP appears with reduced cross-class connectivity in our embeddings (again marked with a black dotted box). This is also observed for the Pavia dataset. This also corroborates our quantitative improvements in accuracy and robustness.

\begin{tcolorbox}[
    colback=gray!5,
    colframe=black!70,
    boxrule=0.6pt,
    left=2mm,
    right=2mm,
    top=2mm,
    bottom=2mm
]
  
\textbf{\ding{102} Larger Minimum Diameter?}  
A wider basin of convergence provides the optimizer greater flexibility and reduces sensitivity to initialization.

\textbf{\ding{102} Slightly Rugged Landscape (\(\nabla^2 L(\theta) \neq 0\))?}  
Local curvature, captured by the Hessian, \(H(\theta) = \nabla^2 L(\theta)\), encodes fine-grained variations. Moderate eigenvalues, \(\lambda_i(H) > 0\) enrich the representation space.

\textbf{\ding{102} Slight slant from numerous directions?} 
If $\nabla L(\theta)\neq 0$, the loss surface is slightly tilted, i.e., there exists at least one direction of descent. 
When the slope is small ($\|\nabla L(\theta)\|\ll 1$), the drift is updated slowly, often biasing the trajectory toward flatter regions of the landscape. 
To analyze this behavior, we consider the second-order Taylor expansion:  $
L(\theta+\delta)\approx L(\theta)+\nabla L(\theta)^{\top}\delta+\tfrac{1}{2}\delta^{\top}H(\theta)\delta,
$  
where $\theta\in\mathbb{R}^{d^P}$ denotes the parameters, $\delta$ the update step, and $H(\theta)=\nabla^2 L(\theta)$ the Hessian. Here, the first-order term $\nabla L(\theta)^{\top}\delta$ captures the immediate slope-driven change, while the quadratic term $\tfrac{1}{2}\delta^{\top}H(\theta)\delta$ encodes the local curvature. 
This clarifies how both gradient and curvature jointly determine the behavior of updates around $\theta$.  

Flatness can be quantified via the Hessian spectrum as:  
\[
\text{Flatness} \sim \tfrac{1}{n}\sum_{i=1}^n \lambda_i(H(\theta)),
\]  
where $\lambda_i$ are eigenvalues and $n=\dim(\theta)$. 
Smaller $\lambda_i$ imply broader valleys: perturbing $\theta$ along those directions changes the loss very little. 
Such flat regions are associated with stability of the solution and robustness to parameter noise.  

Hence, flatter minima typically correspond to lower expected test loss,  
\[
\mathbb{E}_{x\sim\mathcal{D}}[L(\theta)] \downarrow,
\]  
where $\mathcal{D}$ is the data distribution. 
In our case, the landscape exhibits more local slants from multiple directions. This helps in converging toward the global minimum and aligns with an improved generalization performance.

\end{tcolorbox}

\textbf{Additional Benchmarking Against Pixel-Level and Training-Efficient Approaches}. Table \ref{tab:method_backbone_comparison} highlights how recent SOTAs approach hyperspectral image classification from two distinct directions. GAF-NAU \cite{CVPRWGAF} improves pixel-level representation by converting 1D spectral vectors into 2D angular feature maps using Gramian Angular Field encoding, then applying a neighborhood attention U-Net to suppress irrelevant signals and strengthen class discrimination. In contrast, the Forward-Forward Algorithm (FFA) \cite{CVPRWFFA} targets training efficiency, replacing back-propagation with local goodness functions to reduce computational cost and reduce vanishing gradients. Despite these advances, our patch-based approach achieves a higher accuracy on IP and PU, which shows that patch-level approaches can be a promising avenue in hyperspectral imagery.

\begin{table}[H]
\small
\centering
\caption{Comparison of \textbf{descriptive prompt–based VLM against a vision-only model} on IP and PU.}
\label{tab:prompt_vs_vision}
\begin{tabular}{l c c}
\toprule
\textbf{Methodology} & \textbf{IP (OA \%)} & \textbf{PU (OA \%)} \\
\midrule
\rowcolor{bgecolor}
Descriptive prompts (VLM) & \textbf{94.03} & \textbf{97.26} \\
\rowcolor{e5color}
Vision-only (no prompts)  & 91.52 & 96.60 \\
\bottomrule
\end{tabular}
\end{table}

\vspace{-0.2in}
\begin{table}[H]
\small
\centering
\caption{Comparison with \textbf{additional SOTAs} on IP and PU.}
\label{tab:method_backbone_comparison}
\begin{tabular}{l l l c c}
\toprule
\textbf{Methodology}  & \textit{Venue} & \textbf{IP (OA\%, $\kappa$)} & \textbf{PU (OA\%, $\kappa$)} \\
\midrule

\rowcolor{bgecolor}
{GAF-NAU \cite{CVPRWGAF}}       & \textit{{CVPR'22}} & 81.07 / 78.31 & 91.12 / 88.09 \\
\rowcolor{e5color}
{FFA + BP} \cite{CVPRWFFA}  & \textit{{CVPR’24}}    & 73.65 / 69.78 & 92.51 / 90.11 \\
\midrule
\rowcolor{bgecolor}
\textbf{OURS}   &  & \textbf{94.03} / \textbf{93.54} & \textbf{97.26} / \textbf{96.39} \\
\bottomrule
\end{tabular}
\end{table}

\section{Conclusion}
\label{con}

Our proposed vision-language model for hyperspectral image classification demonstrates that integrating descriptive textual prompts that act as anchors to visual embeddings, significantly enhances feature discriminability, class separability, and generalization, even under limited labeled data. Quantitatively, our approach achieves a substantial accuracy improvement over SOTA models on benchmark datasets with very little parameter footprint compared to leading transformer-based models.  This article also highlights the effectiveness of using both hard and semi-hard negatives along with carefully engineered prompts. 

\textbf{Limitations and Future Works}.   We notice that our method has some limitations, like sensitivity to prompt design and careful selection of text descriptors. For future work, we plan to explore automated prompt optimization to further improve classification performance in cross-scene tasks for real-world application areas.

\subsection*{Acknowledgment}
A part of this research has received support from the IEEE
Geoscience and Remote Sensing Society (GRSS) under the
“ProjNET” scheme.

\end{document}